# Recurrent Amplification and Exact Nonexpansiveness Criteria for Query-Driven Erasure

Haozhe Wang

School of Mathematics and Statistics, Ningxia University, Yinchuan 750021, China

**Abstract**

The Query-derived Erase Direction (QED) augments a rank-one linear attention update with a key-orthogonal query component. Under scalar decay, this correction leaves the eigenvalues of each homogeneous state transition unchanged, while the propagation of an initial perturbation depends on the ordered product of these transitions. We construct a periodic token sequence with fixed key and query projections for which every one-step spectral radius is below one but the spectral radius of the period product exceeds one. The same construction yields a sufficient condition for mean-square growth under independent uniform token sampling. For rank-one erasure with channel-wise decay, we derive a necessary and sufficient spectral-norm test for one-step nonexpansiveness that can be evaluated in $O(d)$ time. The test agrees with direct singular value decomposition on all 1,800 operators evaluated. In a counterfactual replay of a selected head of Qwen3.5-0.8B-Base, the terminal gain on a 4,096-token repeated-text sequence increased from 0.99938 to 1.3977 with the feature and checkpoint gate trajectories held fixed.

## 1 Introduction

Autoregressive attention uses a key–value cache for content-based retrieval, with memory requirements that grow with the context length [1]. Linear attention compresses past keys and values into a fixed-size matrix state, which each new token updates and reads [2, 3]. Associations stored under similar keys may therefore interfere. The delta rule corrects the stored association using the prediction residual at the current key [3, 4]. Gated Linear Attention introduces input-dependent forgetting [5], and Gated DeltaNet (GDN) combines scalar decay with a delta-rule update [6].

Kimi Delta Attention applies decay separately to each key channel and retains a shared scalar gate for erasing and writing [7]. Gated DeltaNet-2 (GDN-2) gives the erase and write operations separate channel-wise gates [8]. The Query-derived Erase Direction (QED) augments the erase vector with a query-derived term projected orthogonally to the key, while preserving the rank-one structure of the update [9]. Q-Delta incorporates both key- and query-based prediction errors into the state update [10].

In an input-dependent recurrence, the initial perturbation propagates through an ordered product of token-dependent operators. The spectral radius governs the asymptotic behavior of powers of a fixed matrix, and the spectral norm gives the maximum one-step stretch. Under switching, growth also depends on how the directions at successive steps couple [11–13]. Variational Linear Attention (VLA) studies the state norm and the recurrence Jacobian from an online regularized least-squares perspective [14]. Related work traces how pretrained linear attention writes and retains information [15]. The orthogonal correction in QED leaves the nontrivial eigenvalue of the rank-one erasure matrix unchanged [9]. What does this one-step spectral property imply for the ordered product?

We show, by constructing a periodic sequence with fixed projections and admissible gates, that one-step spectral stability can coexist with amplification of the transition product. The diagonal structure of channel-wise decay and the rank-one structure of erasure yield an exact test for one-step nonexpansiveness. Synthetic controls isolate the role of cross-token geometry, and replays of pretrained features quantify the directional sensitivity of finite-horizon gain.

# 2 Model and Problem Formulation

## 2.1 GDN-2 and the QED update

Let $k_t, q_t \in \mathbb{R}^d$ denote the key and query at token $t$, respectively, and let $v_t \in \mathbb{R}^p$ denote the value, with $\| k_t \|_2 = \| q_t \|_2 = 1$. We put the key dimension first in the state matrix $S_t \in \mathbb{R}^{d \times p}$, with readout $o_t = S_t^{\mathrm{T}} q_t$. Without the normalization denominator, outer-product linear attention takes the form

$$S_t = S_{t-1} + k_t v_t^{\mathrm{T}}, \quad o_t = S_t^{\mathrm{T}} q_t. \tag{1}$$

The delta rule updates the state using the residual at the current key. Adding scalar decay gives the Gated DeltaNet recurrence [6].

$$S_t = \alpha_t \left( I - \beta_t k_t k_t^{\mathrm{T}} \right) S_{t-1} + \beta_t k_t v_t^{\mathrm{T}}, \quad 0 < \alpha_t, \beta_t < 1. \tag{2}$$

For channel-wise decay and erasure, let $D_t = \mathrm{Diag}\left( \delta_{t,1}, \ldots, \delta_{t,d} \right)$ with $\delta_{t,j} = \exp\left( g_{t,j} \right) \in (0,1)$. The erase and write gates are $b_t \in (0,1)^d$ and $w_t \in (0,1)^p$, respectively. GDN-2 uses the gated key as its erase vector. QED adds the key-orthogonal projection of the gated query, as in (3) [8, 9].

$$\begin{aligned} S_t &= \left( I - k_t a_t^{\mathrm{T}} \right) D_t S_{t-1} + k_t \left( w_t \odot v_t \right)^{\mathrm{T}}, \\ a_t &= b_t \odot k_t + \lambda_t \left( I - k_t k_t^{\mathrm{T}} \right) \left( b_t \odot q_t \right), \end{aligned} \tag{3}$$

Here $\odot$ denotes the Hadamard product. In the original QED formulation, each head has a fixed correction strength parameterized by $\lambda_h = 0.25 \ \mathrm{sigmoid}\left( \theta_h \right)$. Our model and one-step test allow the correction strength to vary across tokens. Each periodic construction and counterfactual replay uses a constant strength, with $\lambda = 0.24$ in the stress test. Define $q_t^{\perp} = \left( I - k_t k_t^{\mathrm{T}} \right) q_t$. For scalar decay and a scalar erase gate, $D_t = \alpha_t I$ and $b_t = \beta_t 1$, respectively, (3) reduces to

$$a_t = \beta_t \left( k_t + \lambda_t q_t^{\perp} \right), \ S_t = \alpha_t \left( I - k_t a_t^{\mathrm{T}} \right) S_{t-1} + k_t \left( w_t \odot v_t \right)^{\mathrm{T}}. \tag{4}$$

When $w_t = \beta_t \mathbf{1}$, the write term in (4) reduces to its counterpart in (2).

Figure 1(a) shows decay, erasure, writing, and readout after the update. The erase vector selects the content to remove from the previous state, and the rank-one correction acts along the key direction in state space. The constructions in Section 3 and the replays in Section 5 use the scalar decay and erase gate in (4). Section 4 allows decay to vary across key channels.

## 2.2 Homogeneous dynamics and stability measures

Consider two runs with different initial states and identical tokens, gates, and writes. The value-dependent write term cancels in the difference, which evolves according to the homogeneous recurrence in (5).

$$\Delta S_t = A_t \Delta S_{t-1}, \quad A_t = \left(I - k_t a_t^{\mathrm{T}}\right) D_t. \tag{5}$$

The state transition product from time zero to $T$ is

$$P_T = A_T A_{T-1} \cdots A_1, \quad \Delta S_T = P_T \Delta S_0. \tag{6}$$

The one-step spectral radius $\rho\left(A_t\right)$, one-step spectral norm $\| A_t \|_2 = \sigma_{\max}\left(A_t\right)$, and finite-horizon gain $\| P_T \|_2$ describe different aspects of perturbation propagation. For any initial perturbation, $\| \Delta S_T \|_F \le \| P_T \|_2 \| \Delta S_0 \|_F$. If $\| P_T \|_2 > 1$, some unit-norm initial perturbation is amplified over the horizon. The bound $\rho\left(A_t\right) < 1$ alone does not guarantee nonexpansiveness of the product. In (4), $k_t^{\mathrm{T}} q_t^{\perp} = 0$, so $I - k_t a_t^{\mathrm{T}}$ has eigenvalues $1 - \beta_t$ and 1, the latter with multiplicity $d-1$. Thus

$$\operatorname{spec}(A_t) = \left\{\alpha_t (1-\beta_t), \alpha_t, \ldots, \alpha_t\right\}. \tag{7}$$

The expression in (7) is independent of $\lambda_t$. However, $A_t$ is generally nonnormal, so its largest singular value and the gain of the time-ordered product may vary with $\lambda_t$ and the query direction. For nonscalar $D_t$, the eigenvalue formula in (7) need not hold for the full transition $A_t$.

Figure 1(b) distinguishes the one-step spectrum from the maximum one-step stretch. Figure 1(c) previews the construction in Section 3: two periodic sequences share the same one-step spectral radii and spectral norms, yet one product grows and the other contracts.

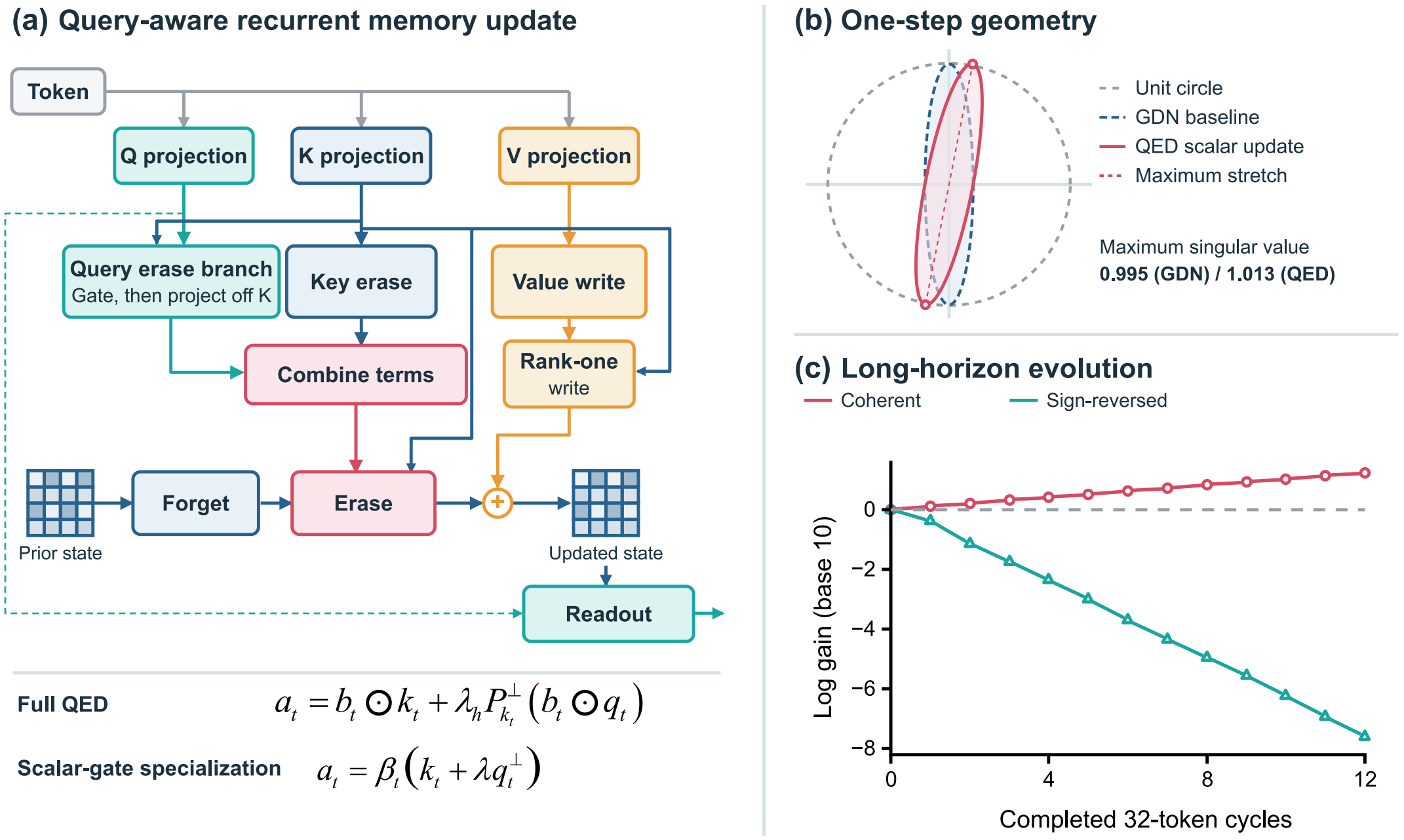

Figure 1. Query-driven erasure and state propagation. (a) Gated decay, rank-one erasure, writing, and post-update readout. (b) Image of the unit circle under the one-step state transition. (c) Prefix gains for two periodic sequences with identical one-step spectral radii and spectral norms.

## 3 Recurrent Amplification under One-Step Spectral Stability

### 3.1 Strict amplification under periodic token sequences

To isolate cross-token coupling, we hold the gates and the angle between the key and query fixed at every step. In dimension $d$ with at least two coordinates, take one-hot tokens $x_i = e_i$ indexed by $i = 1,\ldots,d$. Fix the key projection to $W_K = I$ and set

$$k_i = e_i, \quad q_i = \epsilon e_i - \frac{\sqrt{1-\epsilon^2}}{\sqrt{d-1}} \sum_{j \neq i} e_j, \quad 0 < \epsilon < 1. \tag{8}$$

Each $q_i$ is a column of the fixed matrix $W_Q$, indexed by $i$. Thus the key and query are two fixed linear projections of the same token, and $\| q_i \|_2 = 1$ and $q_i^{\mathrm{T}} k_i = \varepsilon$. We study the difference between states driven by the same tokens, gates, and writes, so the common value-dependent write cancels. Use a constant decay factor $\alpha \in (0,1)$, scalar erase gate $\beta \in (0,1)$, and correction strength $\lambda \in (0, 0.25)$. For token type $i$, write the state transition as $A_i = \alpha R_i$, where $R_i = I - e_i a_i^{\mathrm{T}}$.

**Proposition 1 (Sufficient condition for periodic amplification).** Use the fixed key and query projections in (8) and present all one-hot tokens in order. Suppose that

$$\alpha^d \left[ 1 - \beta + \beta\lambda\sqrt{1-\varepsilon^2}\sqrt{d-1} \right] > 1, \tag{9}$$

Then every one-step state transition satisfies $\rho(A_i) = \alpha < 1$. The product over one complete period in the order $1, 2, \ldots, d$ satisfies

$$\rho(A_d \cdots A_1) \geq \alpha^d \left[ 1 - \beta + \beta\lambda\sqrt{1-\varepsilon^2}\sqrt{d-1} \right] > 1. \tag{10}$$

*Proof.* Let $c = \beta\lambda\sqrt{1-\varepsilon^2} / \sqrt{d-1}$. At the update indexed by $i$, the matrix $R_i$ leaves the other coordinates unchanged and maps coordinate $i$ to

$$(1-\beta)s_i + c\sum_{j \neq i} s_j .$$

The one-hot keys and nonpositive key-orthogonal query components in (8) make the off-diagonal coefficients of $R_i$ nonnegative. Its diagonal coefficients are also nonnegative. Starting from the all-ones vector, condition (9) makes the first updated coordinate greater than one. Each update keeps every coordinate at least one. Starting at 1 and updating each coordinate in order, the newly updated coordinate is at least $r_0 = 1 - \beta + c(d-1) > 1$, so $R_d \cdots R_1 1 \succeq r_0 1$. The spectral-radius lower bound for nonnegative matrices gives $\rho(R_d \cdots R_1) \geq r_0$. Multiplying by the common decay factor $\alpha^d$ yields (10). Equation (7) gives a modulus strictly below one for every one-step eigenvalue. □

### 3.2 Mean-square growth under independent token sampling

With the same parameters, we can replace the periodic order with independent uniform token indices $i_t \sim \mathrm{Unif}\{1,\ldots,d\}$. For the initial state $s_0 = 1$, permutation symmetry gives the following eigenvalue of the mean transition along the all-ones direction:

$$m=\alpha\left[1+\frac{\beta}{d}\left(\lambda\sqrt{1-\varepsilon^2}\sqrt{d-1}-1\right)\right]. \tag{11}$$

It follows that $\mathbb{E}[s_t]=m^t 1$. When $m>1$, the mean state grows exponentially. Jensen's inequality gives $\mathbb{E}\|s_t\|_2^2 \ge \|\mathbb{E}s_t\|_2^2$, so the second moment is unbounded for the all-ones initial direction. This is a sufficient condition for mean-square growth.

To compute the mean-square evolution directly, use the state vector from (11). Independent sampling makes the current transition independent of the previous state, yielding the second-moment recurrence

$$C_t=\mathrm{E}[s_t s_t^{\mathrm{T}}],\quad C_t=\mathcal{L}(C_{t-1}),\quad \mathcal{L}(C)=\frac{1}{d}\sum_{i=1}^{d}A_i C A_i^{\mathrm{T}}.$$

The expected squared state norm is the trace of the second-moment matrix. With the all-ones initial condition, all diagonal entries are equal and all off-diagonal entries are equal. It therefore suffices to propagate one entry of each type. Section 5.1 uses this recurrence to compute the mean-square growth.

## 4 Exact Nonexpansiveness Criteria

The construction in Section 3 shows why one-step spectral radii alone do not control the norm of an ordered transition product. The corresponding one-step norm test asks whether the maximum Euclidean stretch exceeds a prescribed threshold. The spectral norm is at most that threshold exactly when the threshold squared times the identity, minus the Gram matrix of the transition, is positive semidefinite. Diagonal decay and rank-one erasure make this test inexpensive.

### 4.1 Rank-one transitions with channel-wise decay

For the channel-wise decay model $A=\left(I-ka^{\mathrm{T}}\right)D$, let $\|k\|_2=1$ with $D=\mathrm{Diag}\left(\delta_1,\ldots,\delta_d\right)$. Choose the spectral-norm threshold $\eta>\max_j\delta_j$ and define

$$H_\eta=\mathrm{Diag}\left(\eta^2-\delta_j^2\right)+\left(Dk\right)\left(Dk\right)^{\mathrm{T}},\ \ z=D\left(a-k\right). \tag{12}$$

**Theorem 1 (Exact one-step spectral-norm criterion).** Under these assumptions, $H_\eta$ is positive definite because the threshold exceeds the largest diagonal decay factor. Equation (13) defines the test quantity and gives a necessary and sufficient condition for the one-step spectral norm to lie at or below the threshold.

$$\boxed{\|A\|_2\le\eta \Leftrightarrow \Psi_\eta\left(a\right):=z^{\mathrm{T}}H_\eta^{-1}z\le 1.} \tag{13}$$

*Proof.* Using $k^{\mathrm{T}}k=1$, decompose the erasure matrix as $I-ka^{\mathrm{T}}=\left(I-kk^{\mathrm{T}}\right)-k\left(a-k\right)^{\mathrm{T}}$. The output ranges of the two summands are orthogonal, so

$$A^{\mathrm{T}}A=D^2-\left(Dk\right)\left(Dk\right)^{\mathrm{T}}+zz^{\mathrm{T}}. \tag{14}$$

Consequently, $\eta^2 I-A^{\mathrm{T}}A=H_\eta-zz^{\mathrm{T}}$. Since $H_\eta\succ 0$, the rank-one positive semidefiniteness test gives $H_\eta-zz^{\mathrm{T}}\succeq 0$ if and only if $z^{\mathrm{T}}H_\eta^{-1}z\le 1$. This is equivalent to $\|A\|_2\le\eta$. □

Let $r_j = \eta^2 - \delta_j^2$. The Sherman–Morrison identity gives the test quantity as

$$\Psi_\eta(a) = \sum_{j=1}^{d} \frac{\delta_j^2 (a_j - k_j)^2}{r_j} - \frac{\left( \sum_{j=1}^{d} \frac{\delta_j^2 (a_j - k_j) k_j}{r_j} \right)^2}{1 + \sum_{j=1}^{d} \frac{\delta_j^2 k_j^2}{r_j}}. \tag{15}$$

Equation (15) involves weighted sums of length $d$, so it can be evaluated in $O(d)$ time without forming $A^{\mathrm{T}} A$ or computing a full singular value decomposition (SVD). For $a(\lambda) = a_0 + \lambda h$ in (3), $z(\lambda)$ is affine in $\lambda$. The criterion $\Psi_\eta(a(\lambda)) \le 1$ therefore gives a quadratic inequality in the correction strength, which determines the admissible range for a given token and set of gates.

### 4.2 Closed-form scalar-gate criterion and product bounds

When $D = \alpha I$, $a = \beta(k + \lambda q^\perp)$, and $0 < \alpha < \eta$, Theorem 1 reduces to

$$\| A \|_2 \le \eta \Leftrightarrow \frac{\alpha^2 (1-\beta)^2}{\eta^2} + \frac{\alpha^2 \beta^2 \lambda^2 \| q^\perp \|_2^2}{\eta^2 - \alpha^2} \le 1. \tag{16}$$

At $\eta = 1$, (16) gives the exact one-step Euclidean nonexpansiveness condition, including equality. If $\| A_t \|_2 \le \eta$ holds at every step of an input sequence, submultiplicativity of the spectral norm gives

$$\| P_T \|_2 \le \prod_{t=1}^{T} \| A_t \|_2 \le \eta^T. \tag{17}$$

Thus $\eta = 1$ guarantees nonexpansiveness of every prefix, and a uniform choice $\eta < 1$ gives exponential decay of homogeneous state perturbations. For the full recurrence with uniformly bounded writes, $\eta < 1$ also yields a uniform bound on the state norm. Equation (17) provides a sufficient guarantee. A particular sequence can have a nonexpansive product even when some one-step spectral norms exceed one.

## 5 Numerical Results

The synthetic experiments use 32-dimensional normalized keys and queries to study periodic amplification, independent random inputs, and controlled changes in directional structure. We quantify periodic amplification by the spectral radius of a full period. For query readouts, we use the mean squared error (MSE) against a zero target, which we call the zero-target query risk. In feature replays, the largest singular value of the transition product measures the finite-horizon gain of an initial perturbation.

### 5.1 Recurrent amplification on synthetic sequences

We isolated the effect of the key-orthogonal query correction by keeping the key and query projections and the gates fixed. With $d = 32$, $\varepsilon = 0.2$, $\beta = 0.8$, $\alpha = 0.995$, and $\lambda = 0.24$, the analytic lower bound in (10) was 1.062548, and the spectral radius of the period product from (8) was 1.263654. In the baseline with the correction disabled, $\lambda = 0$, it was 0.170360.

Flipping the key-orthogonal query component of every token in (8) preserves key–query alignment, the norm of that component, and all gates. In both sequences, every one-step spectral radius was 0.995 and every one-step spectral norm was 1.013147. The sign flip reduced the period spectral radius to 0.222845. Repeating the sign-flipped period led to contraction, while repeating the original coherent period led to growth (Figure 1(c)).

The common scalar decay compounds across the steps of a period. To determine when it offsets directional coupling, we evaluated the period spectral radius at four decay factors and 50 correction strengths. Figure 2(a) shows the boundary at unit spectral radius. For the parameter setting above, independent uniform token sampling gave a mean growth factor of 1.002693 and a second-moment recurrence with spectral radius 1.005420. Across 4,000 independent random trajectories, the sample mean of the squared state norm showed the same growth trend as the theoretical expectation at the recorded steps.

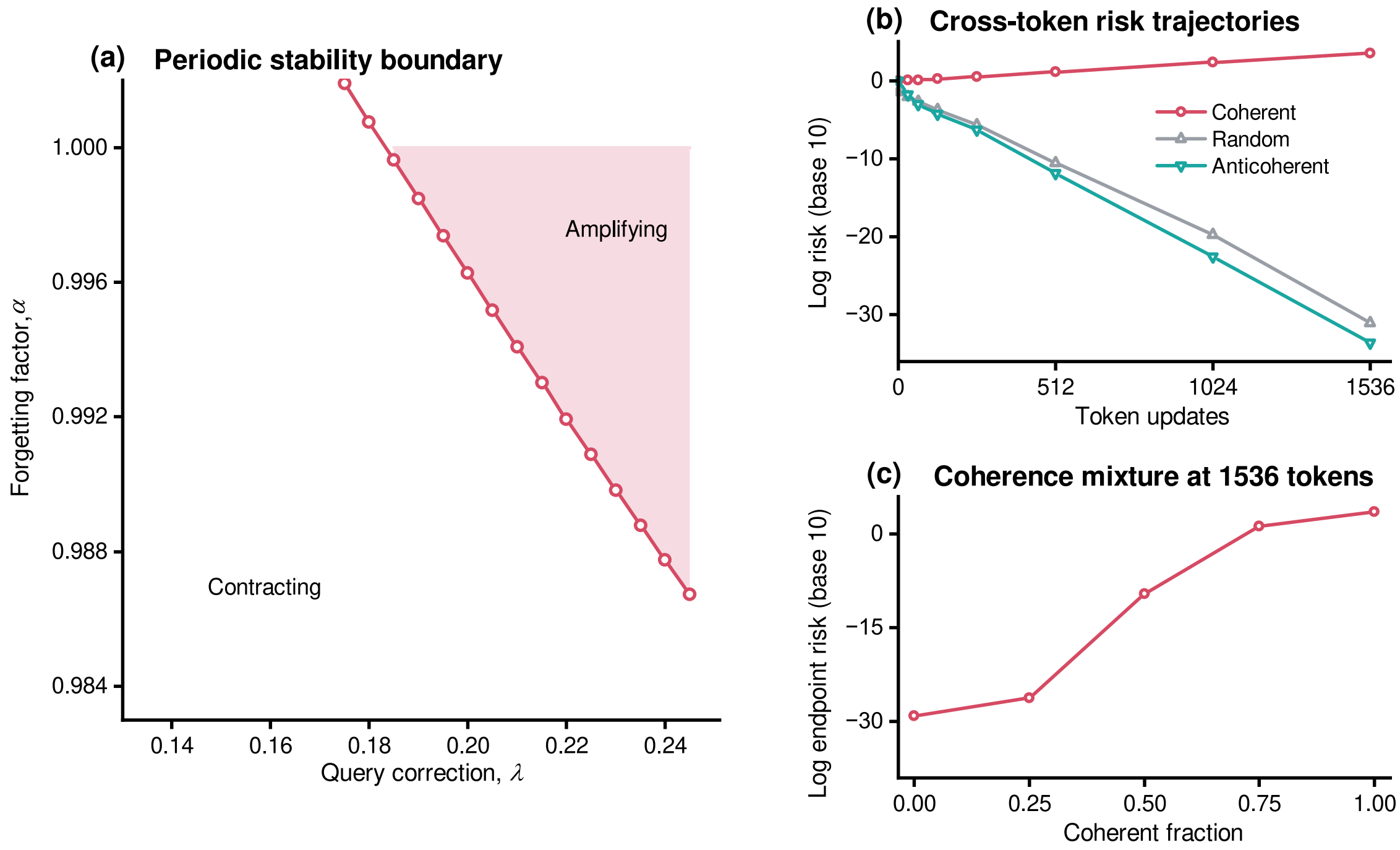

Figure 2. Periodic spectral boundary and zero-target query risk for synthetic sequences. (a) Boundary defined by the spectral radius of the 32-step period product, with $\rho(P_{32})=1$. Line segments connect adjacent samples. (b) Risk trajectories under three arrangements of the key-orthogonal query directions. (c) Terminal risk at step 1,536 versus the coherent fraction. Panels (b) and (c) display the base-10 logarithm of the risk.

### 5.2 Effects of cross-token query geometry

We held $\alpha$, $\beta$, $\lambda$, $q_i^{\mathrm{T}} k_i$, and $\| q_i \|_2$ fixed at each step and varied the arrangement of the key-orthogonal query directions. All three sequences had identical one-step spectral radii and spectral norms. At step 1,536, the risk in the coherent sequence reached approximately 3,913 times its initial value. The risk declined in both the random and anti-coherent sequences.

Figures 2(b) and 2(c) plot the base-10 logarithm of the absolute risk. The 3,913-fold increase is the ratio of terminal to initial risk within the coherent sequence. Across the five mixture fractions in Figure 2(c), higher coherent fractions gave larger terminal risks. Because the one-step quantities were matched, the controlled comparison links these changes to the arrangement of directions across steps. Figure S1(a) shows the negative control with 20 random joint key and query projections.

### 5.3 Numerical verification of the exact criterion

We used (15) and full SVD to test $\|A\|_2 \le 1$ for 1,800 one-step operators generated from random normalized keys and queries, channel-wise decay, and erase gates. The SVD classified 1,565 operators as satisfying the condition and 235 as violating it. The criterion agreed in every case. The maximum absolute difference between the test values from (15) and a direct linear solve was approximately $1.22\times10^{-14}$.

### 5.4 Counterfactual replay of pretrained features

We extracted post-convolution, normalized $q_t, k_t$ and checkpoint gates from the GDN layers of Qwen3.5-0.8B-Base [16]. The exploratory screening produced 576 summary records across three text types (natural, repeated, and topic-alternating), two lengths (2,048 and 4,096 tokens), six layer positions, two heads per layer, two gate settings, and four correction strengths. Head 8 in layer 0 showed terminal amplification on all three text types. We independently verified the repeated-text trajectory for this head. The counterfactual comparison held the keys, queries, and checkpoint gates fixed and used $\lambda = 0$ as the baseline. We use $P_T$ for the homogeneous transition product on a fixed feature trajectory and $\sigma_{\max}\left(P_T\right)$ for its worst-case directional gain.

On the 4,096-token repeated-text sequence, the correction $\lambda = 0.24$ increased the terminal gain from the GDN baseline of 0.99938 to 1.3977. Terminal amplification also occurred for the other two text types (Figure 3(a) and Table 1). Figure 3(b) shows prefix gains at 256-token intervals. The eigenvalues of each one-step transition along the repeated-text trajectory had modulus below one.

Table 1. Terminal gain, forgetting deficit, and adjacent-query alignment for the selected head on 4,096-token sequences

| Text type | GDN baseline terminal gain | QED replay terminal gain | Mean forgetting deficit | Median adjacent cosine |
|---|---|---|---|---|
| Natural | 0.999370 | 1.294446 | $8.026\times10^{-8}$ | 0.802 |
| Repeated | 0.999378 | 1.397709 | $1.136\times10^{-7}$ | 0.813 |
| Topic-alternating | 0.999484 | 1.241276 | $4.922\times10^{-8}$ | 0.819 |

The table uses the checkpoint gates of head 8 in layer 0. The GDN baseline is the homogeneous replay with the query correction disabled. The QED replay uses $\lambda = 0.24$. The forgetting deficit is $1-\alpha_t$. The cosine is the signed cosine between the unit key-orthogonal query components at adjacent tokens.

At $\lambda = 0.24$, the full transition product had 12 singular values greater than one. The largest singular value gives the gain along the most amplified initial direction. The squared gain of a random unit vector averages squared singular values with weights given by its squared components in the corresponding right singular directions. On repeated text, the root-mean-square gain over random directions was 0.58573, and the maximum gain observed among 100,000 probes was 0.79491 (Figure 3(c)).

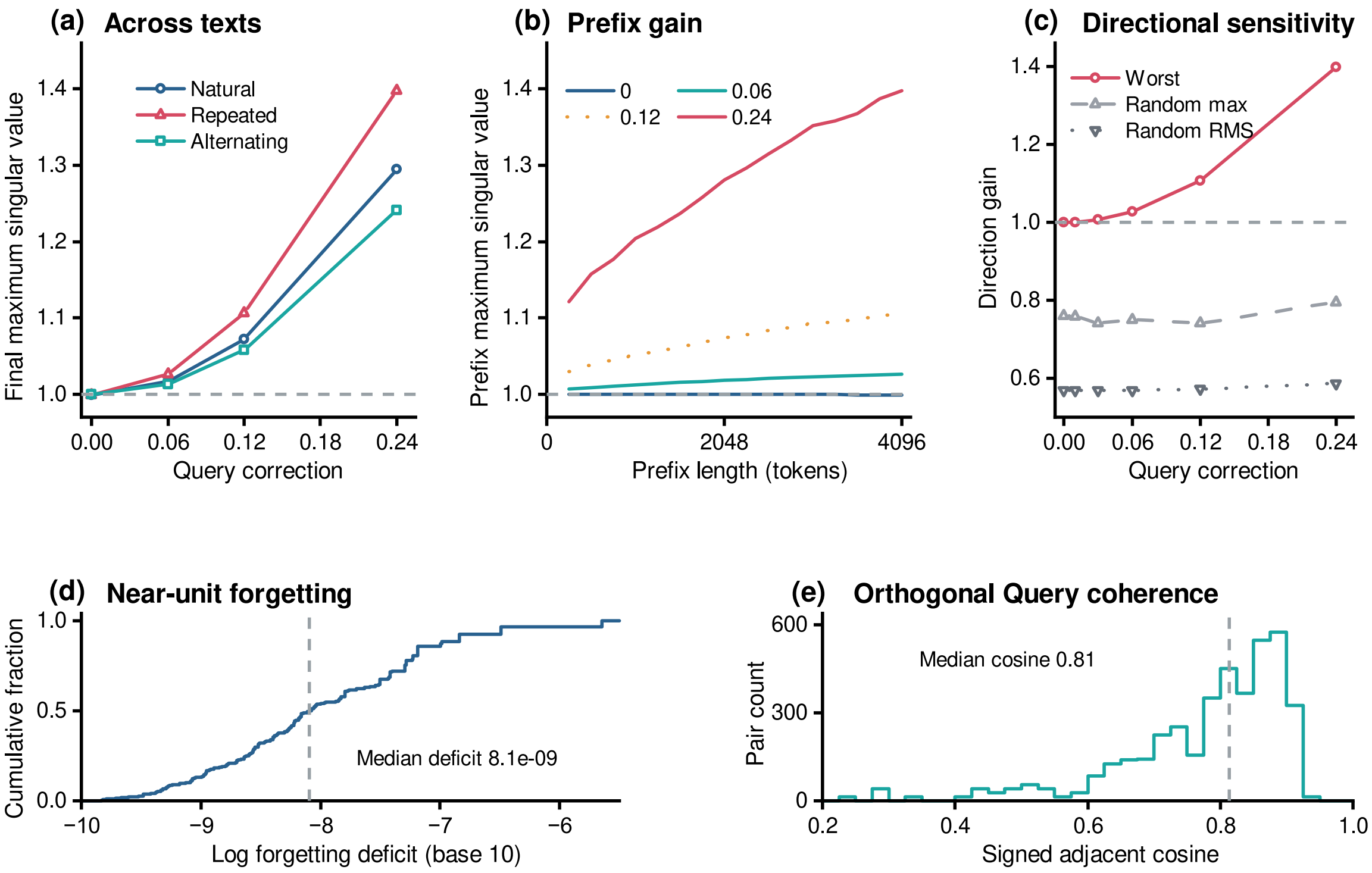

Figure 3. Counterfactual replay of homogeneous QED dynamics using pretrained features. (a) Terminal gains for three text types. (b) Prefix gains on repeated text. (c) Worst-case and random-direction gains. (d) Distribution of the forgetting deficit on repeated text. (e) Distribution of signed cosines between adjacent key-orthogonal query directions on repeated text.

On repeated text, the median forgetting deficit was approximately $8.1\times10^{-9}$, and the median signed cosine between adjacent key-orthogonal query directions was 0.813 (Figures 3(d) and 3(e)). Topic-alternating text had a slightly higher median adjacent cosine but lower terminal gain (Table 1). The adjacent cosine describes pairwise query alignment. The full product also depends on the key directions, the magnitude of the correction, and coupling at longer lags.

The controlled-gate replay kept the features fixed and used $\alpha=0.995$, $\beta=0.8$, and $\lambda=0.24$. Across the 36 trajectories (three text types, six layer positions, and two heads per layer, at 4,096 tokens), every step violated the one-step nonexpansiveness criterion, yet every sampled positive-length prefix was contractive. Expansive directions at one step can become misaligned with high-gain directions at later steps, and decay can further attenuate accumulated perturbations.

## 6 Discussion and Conclusion

The orthogonal projection in QED preserves the nontrivial eigenvalue of the rank-one erasure matrix [9]. The periodic counterexample matches both the one-step spectra and the one-step norms across the two sequences. Their different long-run behavior therefore depends on how successive updates act on the perturbations produced by preceding updates. The criterion controls the largest singular value at each step and thereby bounds every prefix gain. Along a particular trajectory, later updates can offset expansion at an individual step.

The counterfactual replay isolates the effect of changing the erase direction along a trajectory of fixed features and gates from a GDN checkpoint. The zero-correction baseline disables only the query-derived orthogonal erase correction. The strength of 0.24 is close to the QED parameter bound of 0.25 and serves as a stress test. The original QED study reports an initialization near 0.03 [9]. The observed gains apply to the fixed inputs and sampled heads. Evaluating features, gate distributions, and retrieval after QED training requires checkpoints trained with that update.

The periodic construction establishes that query-driven erasure can amplify a recurrent state transition product despite one-step spectral stability. For channel-wise decay, the exact linear-time spectral-norm test guarantees nonexpansiveness of every prefix when satisfied at each step.

## Supplementary Material

The state norm declined for all 20 independent random joint key and query projections. With nonzero values from a fixed linear teacher, the synthetic proxy query MSE after 2,400 steps was 0.19477, 6.31957, and $2.7082\times10^{8}$ for $\lambda = 0, 0.125, 0.24$, respectively (Figure S1). This proxy depends on the write term, queries, and teacher targets.

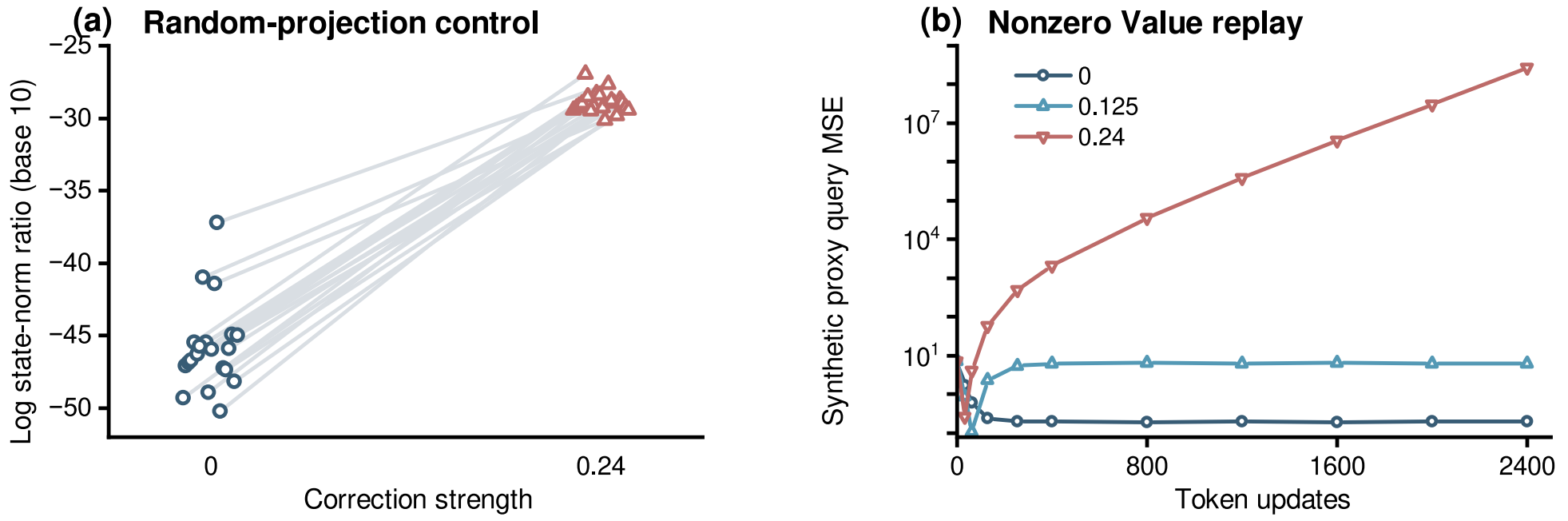

Figure S1. Synthetic controls. (a) Ratio of the final to initial state norm for 20 random key and query projection seeds at two correction strengths, on a base-10 logarithmic scale. (b) Synthetic proxy query MSE with nonzero values.

For the selected head on repeated text, the difference in logarithmic gain between float64 SVD and stepwise replay was below $3\times10^{-15}$. The difference in terminal gain between float32 and float64 was below $1.8\times10^{-5}$. We independently verified the 4,096-token transition product for this head, and treated the remaining heads as exploratory. A 128-token chunked-versus-unchunked comparison gave a maximum key or query RMSE of approximately 0.0086. Table S1 reports the log terminal gains at correction strength 0.24. We factor out scalar decay and accumulate its logarithm, then rescale the geometric product at each step by its Frobenius norm. To obtain the log terminal gain, we add the contributions from decay and rescaling to the logarithm of the largest singular value of the rescaled product and convert the sum to base 10.

Table S1. Log terminal gains for 12 sampled heads under checkpoint gates

| Layer | Head | Natural text | Repeated text | Topic-alternating text |
|---|---|---|---|---|
| 0 | 0 | −7607.8 | −7598.6 | −7820.9 |
| 0 | 8 | 0.11208 | 0.14542 | 0.093869 |
| 1 | 0 | −95.91 | −113.35 | −70.319 |
| 1 | 8 | −1068.7 | −1033.7 | −931.25 |
| 8 | 0 | −524.84 | −187.3 | −393.96 |
| 8 | 8 | −71.336 | −10.788 | −59.942 |
| 9 | 0 | −0.42436 | −0.20676 | −0.44419 |
| 9 | 8 | −1.5896 | −0.48673 | −1.7455 |
| 16 | 0 | −117.92 | −31.075 | −136.9 |
| 16 | 8 | −13.814 | −2.1539 | −16.547 |
| 17 | 0 | −173.31 | −20.022 | −152.82 |
| 17 | 8 | −71.171 | −28.801 | −59.14 |